\documentclass[letterpaper, 10 pt, conference]{ieeeconf}  % Comment this line out if you need a4paper

\IEEEoverridecommandlockouts                              % This command is only needed if
\usepackage{graphics} % for pdf, bitmapped graphics files
\usepackage{epsfig} % for postscript graphics files
\usepackage{times} % assumes new font selection scheme installed
\usepackage{amsmath} % assumes amsmath package installed
\usepackage{amssymb}  % assumes amsmath package installed
\usepackage{multirow}

\usepackage{makecell}
\usepackage{soul}       % for strikethrough \st

\makeatletter
\let\NAT@parse\undefined
\makeatother

\input{preamble.tex}

\definecolor{SaddleBrown}{RGB}{139,69,19}

\definecolor{darkgreen}{rgb}{0,0.5,0}

\title{\LARGE \bf
  Zero-Shot Metric Depth Estimation via Monocular Visual-Inertial
  Rescaling for Autonomous Aerial Navigation
}

\ifthenelse{\equal{\arxivmode}{true}}{
\author{Steven Yang$^*$, Xiaoyu Tian$^*$, Kshitij Goel, and Wennie Tabib%
  \thanks{$^*$ Equal Contributions.}
  \thanks{The authors are with the Robotics Institute, School of Computer Science,
    Carnegie Mellon University, Pittsburgh, PA 15213 USA.
    \{\texttt{yiy6,xiaoyut,kgoel1,wtabib}\}\texttt{@andrew.cmu.edu}.}
}
}{
\author{Author names omitted for anonymous review}
}

\begin{document}

\maketitle
\thispagestyle{empty}
\pagestyle{empty}

%%%%%%%%%%%%%%%%%%%%%%%%%%%%%%%%%%%%%%%%%%%%%%%%%%%%%%%%%%%%%%%%%%%%%%%%%%%%%%%%
\begin{abstract}

  This paper presents a methodology to predict metric depth from monocular RGB
  images and an inertial measurement unit (IMU). To enable collision avoidance
  during autonomous flight, prior works either leverage heavy sensors
  (e.g., LiDARs or stereo cameras) or data-intensive and domain-specific
  fine-tuning of monocular metric depth estimation methods. In contrast, we
  propose several lightweight zero-shot rescaling strategies to obtain metric
  depth from relative depth estimates via the sparse 3D feature map created
  using a visual-inertial navigation system. These strategies are compared for
  their accuracy in diverse simulation environments. The best performing
  approach, which leverages monotonic spline fitting, is deployed in the
  real-world on a compute-constrained quadrotor. We obtain on-board metric depth
  estimates at \SI{15}{\hertz} and demonstrate successful collision avoidance
  after integrating the proposed method with a motion primitives-based planner.

\end{abstract}

%%%%%%%%%%%%%%%%%%%%%%%%%%%%%%%%%%%%%%%%%%%%%%%%%%%%%%%%%%%%%%%%%%%%%%%%%%%%%%%%

\section{INTRODUCTION}
First Person View (FPV) drone pilots leverage a single forward-facing camera
video stream transmitted over a radio feed and sensors embedded in the flight
controller (e.g., IMU) to aggressively maneuver through dense clutter (e.g.,
through tree branches, under bridges, etc.). In contrast, autonomous aerial
systems leverage stereo cameras~\cite{tabib2022tro} or heavy onboard
LiDARs~\cite{tabib2019real} for perception and collision avoidance. The addition
of sensors increases the system's size and mass, which reduces flight time. The
objective of this paper is to demonstrate collision avoidance in flight with the
same set of minimal sensors (a single camera and IMU) used by FPV drone pilots
to navigate in cluttered environments.
\begin{figure}[t]
  \centering
  \subfloat[\label{sfig:glory_shot_camera}]{\includegraphics[width=0.49\linewidth]{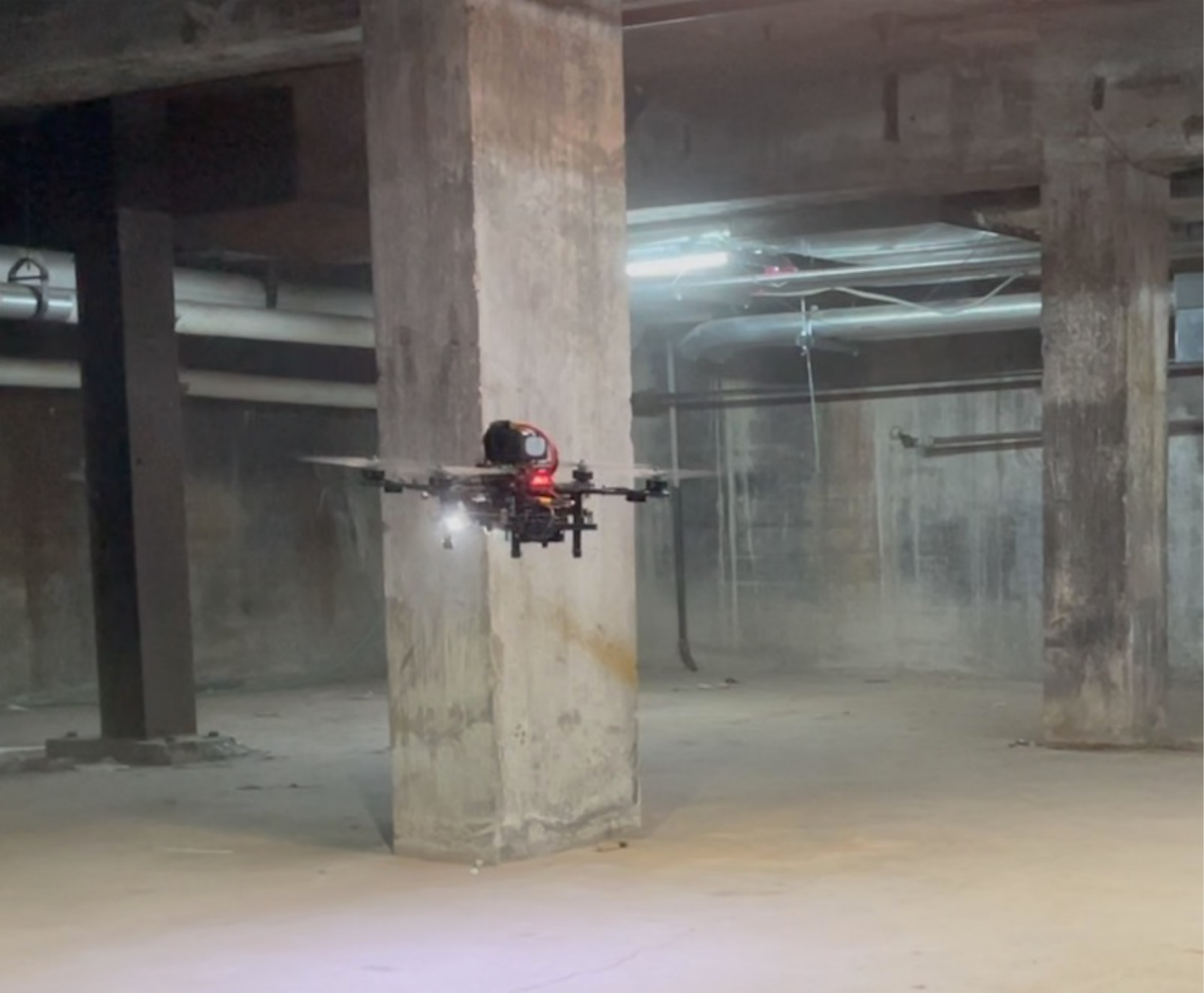}}\hfill
  \subfloat[\label{sfig:glory_shot_faro}]{\includegraphics[width=0.49\linewidth,trim=170 50 100 50,clip]{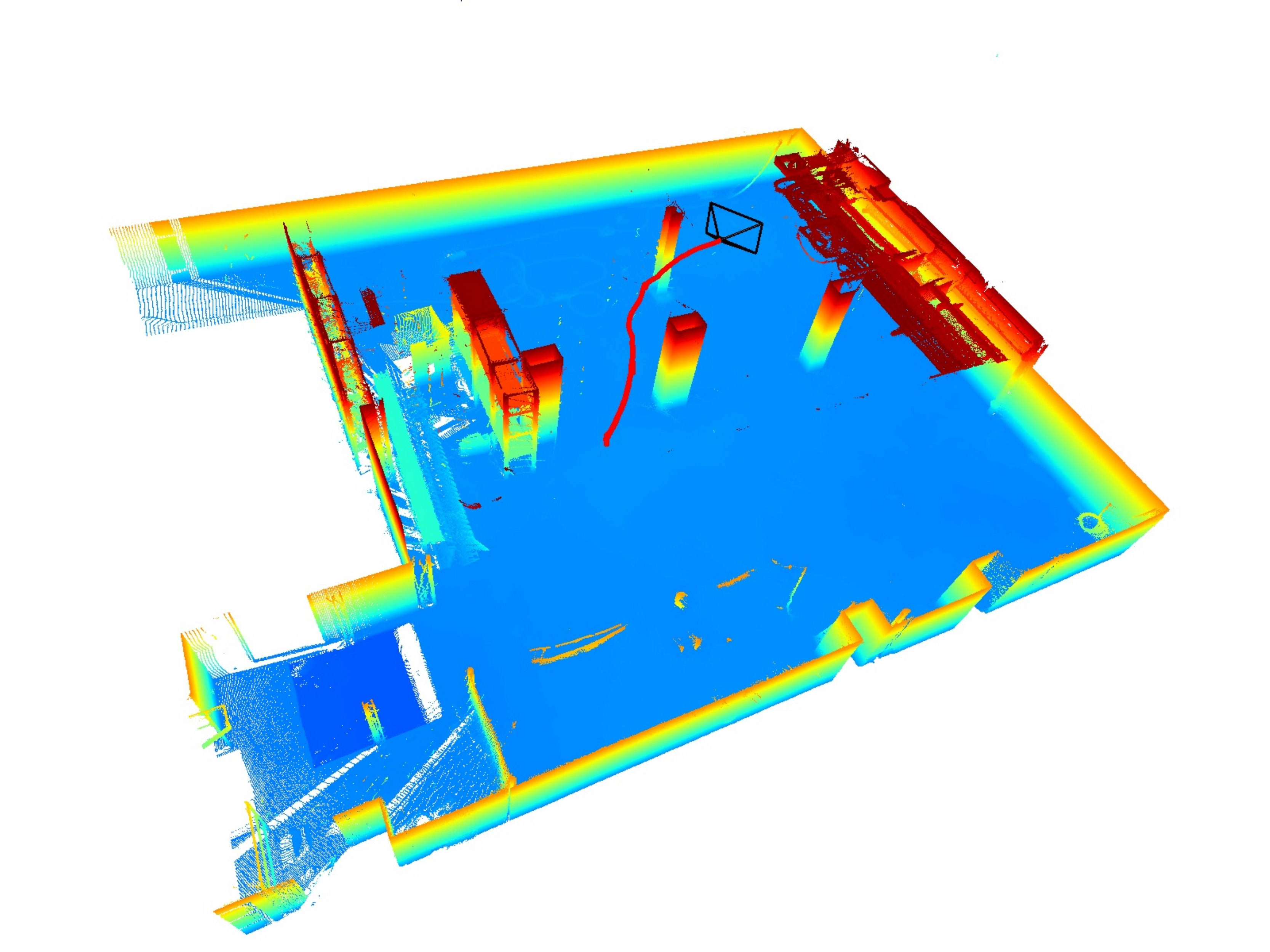}}\\
  \subfloat[\label{sfig:glory_shot_vins}]{\includegraphics[width=0.49\linewidth,trim=10 10 10 0,clip]{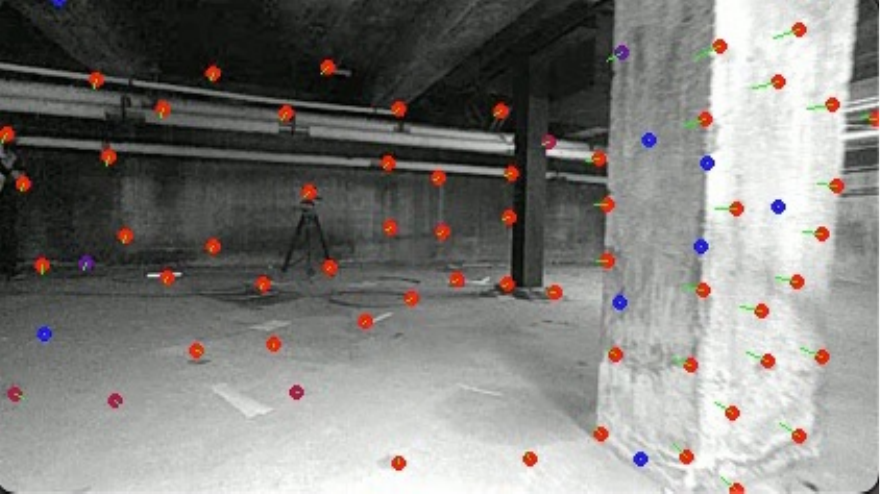}}\hfill%
  \subfloat[\label{sfig:glory_shot_rescaling}]{\includegraphics[width=0.49\linewidth,trim=400 220 200 150,clip]{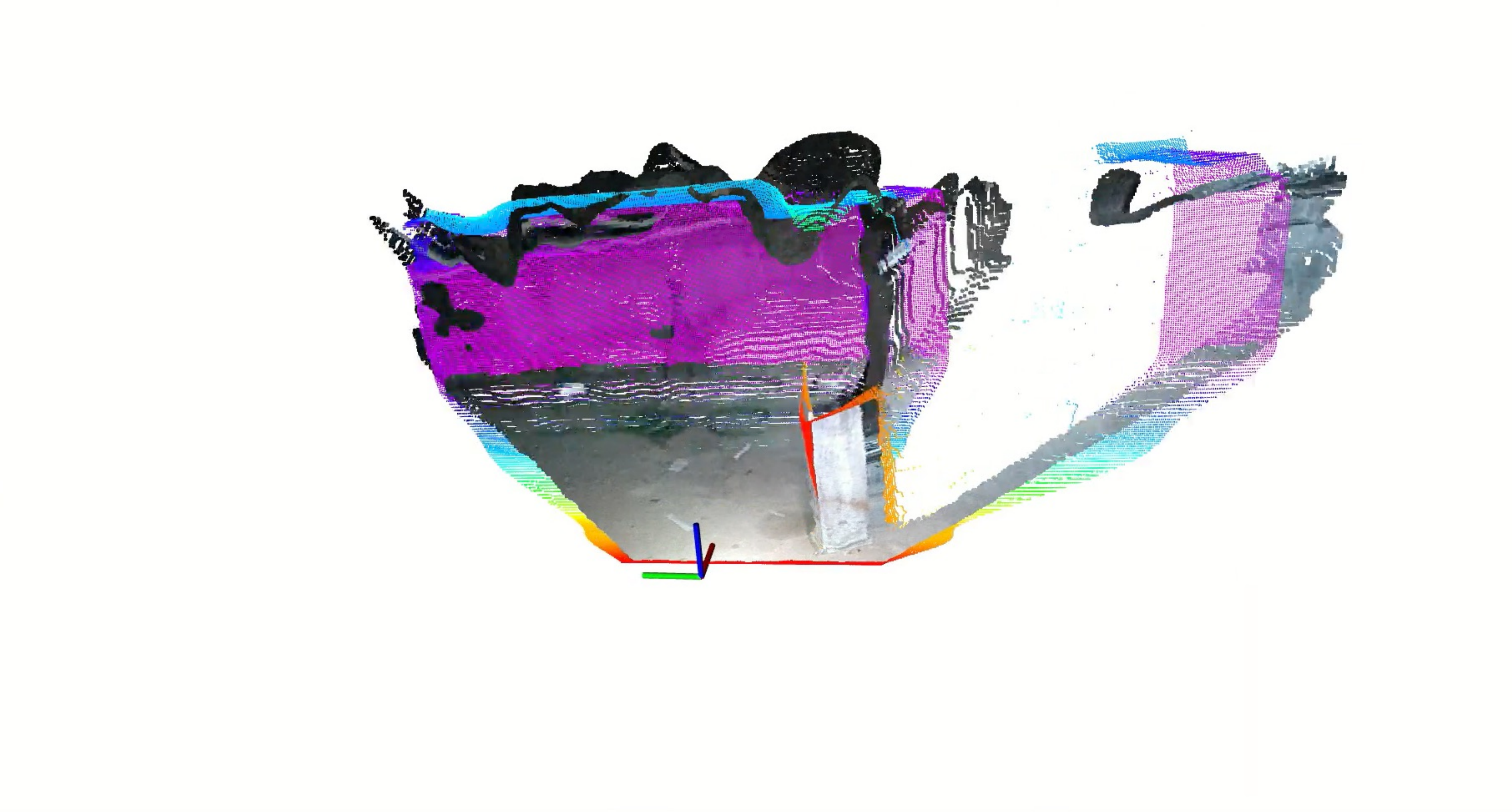}}
  \caption{\label{fig:glory}Image and data corresponding to one hardware experiment
    to demonstrate collision avoidance during autonomous navigation by using
    data from a monocular camera and IMU to rescale relative depth measurements
    from an MDE network and obtain metric depth.
    \protect\subref{sfig:glory_shot_camera} illustrates the quadrotor aerial robot
    navigating in the industrial tunnel environment.
    \protect\subref{sfig:glory_shot_faro} illustrates the trajectory plotted in red
    on top of the environment reconstructed from survey-grade FARO scans.
    This represents the trajectory for the entire flight trial.
    The robot uses the proposed approach to select actions that
    avoid the two pillars in the environment.
    \protect\subref{sfig:glory_shot_vins} shows the features tracked in
    the forward-facing camera, which are used to rescale the predicted image.
    \protect\subref{sfig:glory_shot_rescaling} plots the point cloud generated using
    our approach in colors ranging from red (closer) to purple (further away)
    as well as the colorized point cloud from a RealSense sensor generated using
    active stereo.}
  \vspace{-0.3cm}
\end{figure}

Prior works leverage monocular depth prediction for
aerial navigation via interpolating methods such as the plane sweeping
algorithm~\citep{yang_realtime_2017} to enable motion planning with a single RGB
camera and an IMU~\citep{lin_autonomous_2018}. While successful in large open
environments, the interpolation does not provide sufficient accuracy to avoid thin
obstacles in cluttered environments. Learning-based metrically-accurate
monocular depth estimation (MDE) is a promising
alternative~\citep{piccinelli2025unidepthv2,yang2024depthv2}. However, these
zero-shot approaches require retraining with domain-specific images for
metrically-accurate results. Robots operating in \emph{a priori} unknown environments
(e.g., search and rescue robots) do not have access to data beforehand,
so retraining is challenging for these applications.

Alternatively, recent works correct or complete the depth data reported by an
RGB-D sensor (e.g., RealSense\footnote{
\ifthenelse{\equal{\arxivmode}{false}}{\nolinkurl{http://realsenseai.com}}{\url{http://realsenseai.com}}}) using monocular
depth estimation (MDE) neural networks and rescaling operations for metric
accuracy~\citep{saviolo2024reactive,mao2025time}. However, we are interested in
using one RGB camera and an IMU instead of an active stereo depth sensor.  Prior
zero-shot methods build upon relative MDE methods and propose visual-inertial
fine-tuning~\citep{wofk_monocular_2023} or sparse feature depth test-time
adaptation~\citep{marsal2024simple} for metrically accurate depth images.
However, the former method requires re-training the MDE networks that may lead
to loss of generalization and the latter is evaluated using ground
truth data as input.

To bridge these gaps in the state of the art, we contribute (1) a zero-shot metric depth
estimation method that leverages the sparse 3D feature map from a
visual-inertial navigation system (VINS) for rescaling predicted relative depth
to metric depth, (2) analysis of the proposed rescaling techniques in
challenging simulation environments, (3) hardware
validation of the proposed approach in challenging,
real-world environments onboard a size, weight, and power (SWaP)-constrained
quadrotor aerial system\ifthenelse{\equal{\arxivmode}{true}}{\footnote{A video
of the experiments may be found at~\url{https://youtu.be/t6FajgB06Vc}}}{}
(\cref{fig:glory}), and (4) open-source software release of the rescaling
method\ifthenelse{\equal{\arxivmode}{false}}{\footnote{A link will be added here
upon acceptance of this
paper.}}{\footnote{\url{https://github.com/rislab/mono_depth_rescaler}}}.

\begin{figure*}
  \includegraphics[width=\linewidth,trim=80 720 190 290,clip]{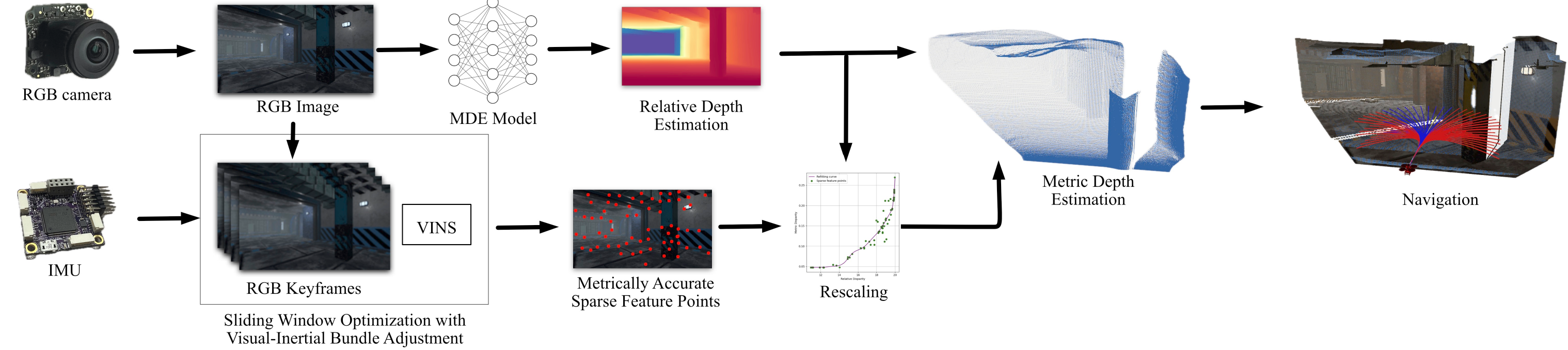}
  \caption{Overview of the approach to rescale predicted depth from an MDE network
      using a metrically accurate 3D sparse feature map from a VIN system.
      The RGB camera image is used by an MDE network to predict a depth image consisting of
      relative depth estimates. The RGB camera images and IMU data are also used to
      produce a sparse set of metrically accurate 3D features. We leverage a monotonic
      spline to rescale the relative depth estimates so that they are metrically accurate.
      The resulting rescaled depth image is used for navigation.
    }
    \vspace{-0.25cm}
\end{figure*}

\section{RELATED WORK}
Zero-shot methods that output metric depth given a monocular RGB image
include ZoeDepth~\citep{bhat2023zoedepth},
UniDepth~\citep{piccinelli2024unidepth,piccinelli2025unidepthv2},
Metric3D~\citep{yin2023metric3d,hu_metric3d_2024}, and Depth
Anything~\citep{yang2024depth,yang2024depthv2}. ZoeDepth requires expensive
fine-tuning for metric depth estimation if the camera calibration at test-time
is different from the one used for training. UniDepth attempts to
learn the camera calibration during training for enhanced generalization.
However, camera calibration data is required at training time, which means that
UniDepth is unable to leverage large datasets of uncalibrated images. Metric3D
is able to introduce invariance to camera calibration at the cost of high
test-time compute requirements. Depth Anything can leverage uncalibrated image
datasets but recommends fine-tuning for high metric accuracy on domain-specific
data for downstream applications. However, for many real-world robotics tasks,
the environmental conditions are \emph{a priori} unknown (e.g., search and rescue).
Therefore, we present an approach that provides
zero-shot metrically-accurate depth estimates from learning-based relative depth
predictions without expensive environment- or camera-specific fine-tuning.

~\citet{wofk_monocular_2023} present a post-training approach to resolve the
scale ambiguity in relative depth estimates and obtain metric depth.
~\citet{marsal2024simple}, which is most similar to our work, rescale the affine-invariant depth
estimates output from an MDE (i.e., Depth Anything~\cite{yang2024depth,yang2024depthv2}). Given
a sparse set of 3D points, the approach performs a linear regression to estimate
two parameters for rescaling the depth estimates.
RANSAC~\cite{fischler1981random} is used to ensure robustness in the regression
solution. However, the sparse set
of 3D points is generated using ground truth data instead of real sensor data,
which may be noisy. The LiDAR sensor is simulated by sampling points along a line in
ground truth depth images. To simulate SFM points, the authors match features
(e.g., SIFT~\cite{Lowe2004}) across consecutive image frames and triangulate the
3D feature locations using the ground truth poses between the images. In both
cases, the sparse 3D feature sets are very accurate because they leverage ground
truth information, which is not representative of real-world operation. In
contrast, our approach relies on the sparse 3D feature points estimated from a
sliding window optimizer~\cite{Yao-2020-120702}. We also do not need extra
onboard sensors like heavy LiDARs or hardware triggered stereo cameras.
Moreover, the evaluation in~\citep{marsal2024simple} is conducted only on
publicly available datasets in postprocessing whereas we deploy our approach to
a SWaP-constrained aerial robot.

Specific to aerial navigation,~\citet{saviolo2024reactive} enhance the depth
estimation of the Intel Realsense D435i by feeding the color camera images into
DepthAnythingV2~\cite{yang2024depth} and using the output to fill holes in the
depth image created from the time-synchronized IR stereo cameras. The monocular
depth estimation network estimates depth but the values are not
guaranteed to be correct up-to-scale~\citep{mao2025time}. To correct for these
scale inaccuracies, the authors fit a quadratic polynomial to the known depth in
the active stereo depth image and corresponding depth values in the learnt depth
image. This polynomial is used to rescale the depth values for the learnt depth
image and fill holes in the active stereo depth completion. In contrast, we
provide an approach to use only one RGB camera instead of a stereo camera setup
and use higher-fidelity but lightweight rescaling methods (e.g., exponential,
monotonic splines) for metric accuracy.

%%% Local Variables:
%%% mode: latex
%%% TeX-master: "../main"
%%% End:

\section{METHODOLOGY}
\subsection{Sparse Feature Depth Map Generation}
This section describes how we derive sparse depth
maps. We use the monocular
visual-inertial navigation system detailed in~\cite{Yao-2020-120702}
with minor modifications.  We summarize the approach and detail the
modifications for our application in the following
paragraphs.

Given a monocular RGB image, Shi-Tomasi corners~\cite{shi1994good} are detected and a
minimum distance is enforced between nearby features to enable salient features
to be tracked across the image frame. The pyramidal Lucas-Kanade
method~\cite{lucas1981iterative} is used to find feature correspondences between
consecutive frames and outliers are rejected using
RANSAC~\cite{fischler1981random,kneip2011robust}.
The IMU motion, biases, and feature locations are optimized by minimizing
a nonlinear objective function over a sliding window of image keyframes that
encodes both IMU- and vision-derived motion constraints.  This optimizer
provides a sparse 3D feature map with respect to the earliest keyframe camera pose
at a rate slower than the camera framerate. To obtain 3D feature depths in the
current camera frame, we project the feature map using the high-rate odometry
generated via IMU forward-propagation (also referred to as upsampled odometry). Formally, if the earliest keyframe
is $C_i$ and the current camera frame is $C_j$, we can calculate each feature position $_{C_j}\mathbf{p}$
from $_{C_i}\mathbf{p}$ through the transformations
\begin{align*}
    _W\mathbf{p}     & = \mathbf{R}_{WB_i} \left( \mathbf{R}_{B_i C_i} \, _{C_i}\mathbf{p} + \mathbf{t}_{B_iC_i} \right) + \mathbf{t}_{WB_i}         \\
    _{C_j}\mathbf{p} & = \mathbf{R}_{B_jC_j}^{-1} \left( (\mathbf{R}_{WB_j})^{-1} ( _W\mathbf{p} - \mathbf{t}_{WB_j}) - \mathbf{t}_{B_jC_j} \right),
\end{align*}
where $B$ and $W$ denote the body and world frames, respectively,
$\mathbf{R}_{PQ}$ denotes the rotation matrix and $\mathbf{t}_{PQ}$ the
translation vector from a frame $P$ to a frame $Q$. The quantities
$\mathbf{R}_{WB_i}$, $\mathbf{R}_{B_i C_i}$, $\mathbf{t}_{B_iC_i}$,
$\mathbf{t}_{WB_i}$, $\mathbf{R}_{B_jC_j}$, $\mathbf{R}_{WB_j}$,
$\mathbf{t}_{WB_j}$, and $\mathbf{t}_{B_jC_j}$ are known via the upsampled
odometry and extrinsic calibration.

With these temporally aligned feature positions, we obtain the
corresponding pixel locations in the image plane using the pinhole camera model.
For fractional locations, we choose the surrounding four integer locations
and select the smallest relative depth location which represents the nearest pixel
to the robot. This is to avoid predicting an obstacle to be further away than it
actually is.

% \[
% u = f_x \frac{X}{Z} + c_x,
% \quad
% v = f_y \frac{Y}{Z} + c_y.
% \]

% where $f_x$, $f_y$ are the focal lengths along the $x$ and $y$ axis, respectively,
% and $c_x, c_y$ is the principal point. We let $p_{c_k} = (X, Y, Z)^\top, \;Z > 0$.
% The depth value for the pixel is given by $d = Z$.

\subsection{Depth Rescaling\label{ssec:depth_fitting}}
Given an RGB image, let $z_{\text{rel}}(i)$ and
$z_{\text{met}}(i)$ be the estimated relative and metric depth values at a pixel
$i$ from the learnt depth model and the sparse feature map, respectively.
Assuming $z_{\text{rel}}(i),z_{\text{met}}(i) > 0$, we can formulate the depth
rescaling problem in terms of disparities $d_{\text{rel}}(i) =
    \frac{1}{z_{\text{rel}}(i)}$ and $d_{\text{met}}(i) =
    \frac{1}{z_{\text{met}}(i)}$~\cite{yang2024depth}. If there are $N$ pixels where the disparity values
$(d_{\text{rel}}(i),d_{\text{met}}(i))\, \forall i \in \{ 1,\dots,N \}$ are
valid, the depth rescaling objective is to derive a scalar-valued function $f$
that maps the relative disparity to metric disparity.  This section details
rescaling methodologies that leverage several forms of $f$: polynomial,
exponential, smoothing (cubic) splines, monotonic smoothing splines, and
monotonic splines.

\subsubsection{Polynomial}
If $f$ is a $n$-th degree polynomial parameterized by the coefficient vector
$\mathbf{a}$, we solve for $\mathbf{a}$ via
\begin{equation}
    \min_{\mathbf{a}} \sum_{i}^{N} \left( f(d_{\text{rel}}(i)) - d_{\text{met}}(i) \right)^{2}.
    \label{eq:poly}
\end{equation}
The solution to this least-squares problem is given by $$ \mathbf{a} =
    (\mathbf{X}^{\top}\mathbf{X})^{-1}\mathbf{X}^{\top}\mathbf{d}_{\text{met}}, $$
where $\mathbf{d}_{\text{met}} =
    [d_{\text{met}}(1),\dots,d_{\text{met}}(N)]^{\top}$ and  $\mathbf{X}$ is the $N
    \times (n+1)$ Vandermonde matrix for $\mathbf{d}_{\text{rel}} =
    [d_{\text{rel}}(1),\dots,d_{\text{rel}}(N)]^{\top}$.

\subsubsection{Exponential}
Let $f$ be an exponential function in the form $f(d_{\text{rel}}) = a\cdot
    e^{bd_{\text{rel}}}$.  If $g(d_{\text{rel}}) = \ln f(d_{\text{rel}}) = \ln a + b
    \cdot d_{\text{rel}} = a' + b \cdot d_{\text{rel}}$, we can use the ordinary
least-squares from~\cref{eq:poly} using $g$:
\begin{equation}
    \min_{a', b} \sum_{i}^{N} \left( g(d_{\text{rel}}(i)) - \ln d_{\text{met}}(i) \right)^{2}.
    \label{eq:exp}
\end{equation}
Finally, we get the original coefficient $a$ using $a = e^{a'}$.

\subsubsection{Smoothing Spline}
We utilize the cubic $C^2$ spline function fitting algorithm derived from~\citet{dierckx1975algorithm}.
The details are referenced here for completeness.
For a cubic $C^2$ spline function $f$, we start with evenly-spaced
and sorted knots $k_j$ for $j = \{4, 5, \dots, n-3\}$ defined along domain $[k_4,
            k_{n-3}]$ such that: (1) Between each interval $(k_j, k_{j+1}),\; j= 4, \ldots,
    n-4$, $f$ is defined as some polynomial with degree $3$ or less and (2) $f$ is
$C^2$ continuous across $[k_4, k_{n-3}]$. Using this
information, we want to minimize the following criterion:
\begin{equation*}
    \sum_{r=5}^{n-4} p_r^2
\end{equation*}
subject to the constraint
\begin{equation*}
    \sum_{i=1}^{N} \left( d_{\text{met}}(i) - f(d_{\text{rel}}(i)) \right)^2 \leq S,
\end{equation*}
where $p_r$ is the third derivative jump discontinuity of $f$ at knot position
$k_r$, defined explicitly as $p_r = f^{(3)}(k_r^+) - f^{(3)}(k_r^-)$ that
represents smoothness, and $S \ge 0$ is a smoothing hyperparameter that controls
the trade-off between minimizing error and maximizing smoothness. A smaller $S$
represents more interpolation towards the data points, while a larger $S$
emphasizes smoothness.

\subsubsection{Monotonic Smoothing Splines}
We leverage the unimodal smoothing formulation from~\citep{eilers2005unimodal}
in this section to enforce monotonicity and smoothness.
Consider $m$ knots $k_1, \ldots, k_m$. Let the cubic (degree $t=3$) B-spline basis functions recursively defined by \citet{de1978practical} as
\[
    \phi_{i,0}(d_{\text{rel}}) =
    \begin{cases}
        1, & k_i \leq d_{\text{rel}} < k_{i+1} \\
        0, & \text{otherwise}
    \end{cases}
    \qquad 1 \leq i \leq m-1.
\]
For degree $t \geq 1$,
\begin{align*}
    \phi_{i,t}(d_{\text{rel}}) = & \frac{d_{\text{rel}} - k_i}{k_{i+t} - k_i}\,\phi_{i,t-1}(d_{\text{rel}})
    \\ & + \frac{k_{i+t+1} - d_{\text{rel}}}{k_{i+t+1} - k_{i+1}}\, \phi_{i+1,t-1}(d_{\text{rel}}),
\end{align*}
where $1 \le i \le m-t-1$.
Let $\beta_j$ be the basis coefficients.
Any cubic B-spline function can thus be written as
\[
    f(d_{\text{rel}}) = \sum_{j=1}^m \beta_j \,\phi_{j,3}(d_{\text{rel}})
\]
where $\phi_{j,3}$ denotes the cubic basis functions for the $j$th knot.

To derive the function, we need to find the $\beta_j$. We solve this through
penalized least squares by minimizing
\[
    \min_{\boldsymbol{\beta}} \; \|\mathbf{d}_{\text{met}} - \mathbf{B}\boldsymbol{\beta}\|^2
    + \lambda \| \mathbf{D}^{(3)} \boldsymbol{\beta} \|^2
    + \kappa \| \mathbf{V}^{1/2} \mathbf{D}^{(1)} \boldsymbol{\beta} \|^2 \\
\]
where $\mathbf{B}$ is the basis matrix with entries $B_{ij} =
    \phi_j(d_{\text{rel}}(i))$, $\mathbf{D}^{(k)}$ is the finite-difference matrix
for the $k$th derivative estimation, $\mathbf{V}$ is a diagonal matrix where
$V_{kk} = 1$ if $\mathbf{D}^{(1)} \boldsymbol{\beta} < 0$ and $0$ otherwise,
$\lambda$ can be used to prioritize smoothing, and $\kappa$ is the non-monotonicity
penalty.

The first term of this objective ensures that $f$ is a \textit{fitting spline}. The second term
enables smoothness by minimizing the third derivative of $f$ (similar to minimizing jerk).
The third term biases $f$ towards monotonicity.

\subsubsection{Monotonic Splines}
This formulation is the same as the preceeding section, but we remove the smoothness term:
\[
    \min_{\boldsymbol{\beta}} \; \|\mathbf{d}_{\text{met}} - \mathbf{B}\boldsymbol{\beta}\|^2
    + \kappa \| \mathbf{V}^{1/2} \mathbf{D}^{(1)} \boldsymbol{\beta} \|^2.
\]

\subsection{Implementation Details}
Depth values outside the range of \SI{0.05}{\meter} to \SI{65}{\meter} are clipped,
only counting as valid the pixels that have ground truth depth within that range
for each frame. The upper bound is set to \SI{65}{\meter} assuming the $16$-bit
unsigned integer encoding for depth is represented in millimeters. The maximal
representable value without overflow is \SI{65}{\meter}, which is also far
enough to support analysis for high-speed navigation. The \SI{0.05}{\meter}
lower bound excludes invalid or zero measurements.

The sample size, or the number of features, plays an important role in
determining the accuracy of the final fitting curve. The required number of
sparse features varies depending on the rescaling method. In general, we require
that there be at least 10 sparse features, which is empirically determined. If
VINS does not provide enough sparse features, we skip the frame with no results
being produced. For all the datasets, this occurs only when VINS is initializing
in the first frame. Furthermore, the number of knots for monotonic smoothing and
monotonic spline rescaling is a hyperparameter that can be tuned for better
results depending on the model and constraints. For our experiments, we find
that 10 knots works well.

%%% Local Variables:
%%% mode: latex
%%% TeX-master: "../main"
%%% End:

\section{EXPERIMENTAL DESIGN AND RESULTS}
\subsection{Simulation Datasets}
\begin{figure}
  \subfloat[Mine\label{sfig:mine}]{\includegraphics[width=0.32\linewidth]{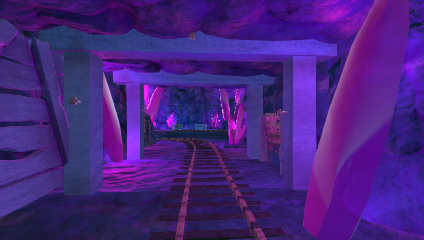}} \hfill%
  \subfloat[Sewer\label{sfig:sewer}]{\includegraphics[width=0.32\linewidth]{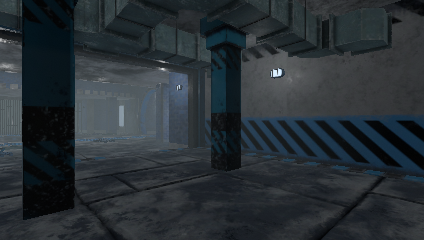}} \hfill%
  \subfloat[Drone Dome\label{sfig:drone_dome}]{\includegraphics[width=0.32\linewidth]{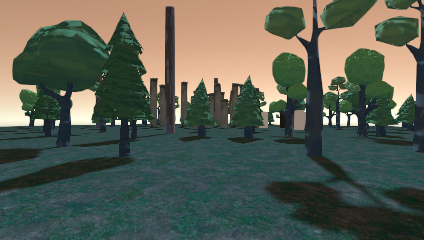}}
  \caption{\label{fig:flightmare_envs}Examples of images used for ablation study derived from photo-realistic
    Flightmare~\citep{song2021flightmare} simulator.
    The environments strike a balance between confined spaces
    (see~\protect\subref{sfig:mine}--\protect\subref{sfig:sewer}) and open
    spaces where the sky may be seen at a distance
    (see~\protect\subref{sfig:drone_dome}).}
  \vspace{-0.25cm}
\end{figure}

To benchmark the rescaling methods with ground truth depth, we collected
datasets from three photo-realistic~\cite{song2021flightmare}
environments: \textit{mine}, \textit{sewer}, and \textit{drone dome}
(\cref{fig:flightmare_envs}). The \textit{mine} contains small confined hallways
with sharp turns that open into a large cavern; the \textit{sewer} environment
contains two large concrete rooms connected with a large hallway; the
\textit{drone dome} is an outdoor cluttered environment with a small forest of
trees and pillars. These environments represent both confined and open
environments where quadrotor aerial robots may be expected to operate.  The
datasets contain VINS system keyframes with associated 3D features. The
evaluation datasets include $846$ frames and 3D keypoint pairs from the
\textit{mine} environment, $707$ from the \textit{sewer} environment, and
$1015$ from the \textit{drone dome} environment.

\subsection{Evaluation Metrics}
%\href{https://discuss.pytorch.org/t/what-does-1-25-1-25-1-25-delta-1-25-stand-for/174841}{HERE}}
%\wtabib{@Steven: The UniDepth paper defines $\delta_i$ as ``the
%percentage of inlier pixels with threshold
%$1.25^i$~\cite{piccinelli2024unidepth}''. I think this may be
%sufficient. We should reword and retain the citation.}
We compare the rescaling approaches detailed in~\cref{ssec:depth_fitting} using
two measures: the absolute relative depth error and $\delta_1$
error~\cite{eigen2014depth}. Let $N$ be the number of valid pixels,
$z_{\text{pred}}(i)$ be the predicted depth and $z_{\text{gt}}(i)$ be the ground
truth depth for the $i$th pixel. The Absolute Relative Error (AbsRel) is defined
in~\cref{eq:absrel} and measures how far the predicted value is from the ground
truth depth normalized by the ground truth depth. These values are summed and
divided by $N$ to provide an average over all predictions. The $\delta_1$
(defined in~\cref{eq:delta1})~\cite{marsal2024simple} measures the proportion of
predictions within 25\% of the ground truth depth.
\begin{align}
  \text{AbsRel} = \dfrac{1}{N} \cdot \sum^{N}_{i = 1} \dfrac{|z_{\text{pred}}(i) - z_{\text{gt}}(i)|}{z_{\text{pred}}(i)}\label{eq:absrel}
\end{align}

\begin{align}
  \delta_1 = \dfrac{1}{N} \cdot \left| \left\{ i ~:~
  \frac{z_{\text{pred}}(i)}{z_{\text{gt}}(i)} < 1.25 , \
  \frac{z_{\text{gt}}(i)}{z_{\text{pred}}(i)}
  < 1.25 \right\} \right| \label{eq:delta1}
\end{align}

\subsection{Rescaling Algorithm Comparison\label{ssec:refitting}}
% depthany fitting method comparison
\begin{table*}
  \setlength{\arrayrulewidth}{0.7pt}
  \renewcommand{\arraystretch}{1.3}
  \setlength{\tabcolsep}{2pt}
  \centering
  \begin{tabular}{|>{\bfseries}p{1.5cm}|c|c||c|c||c|c||c|c||c|c||c|c||c|c||c|c|}
    \hline\multicolumn{1}{|c|}{\textbf{\makecell{Fitting                                                                                                                                                                                                                                                                                                                                                                                                                                                                                                                                                                                                                                                                                                                                      \\Technique}}} & \multicolumn{4}{c||}{\textbf{Mine}} & \multicolumn{4}{c||}{\textbf{Sewer}} & \multicolumn{4}{c||}{\textbf{Drone Dome}} & \multicolumn{4}{c|}{\textbf{Weighted Average}} \\
    \cline{2-17}
    \multicolumn{1}{|c|}{}     & \multicolumn{2}{c||}{\textbf{AbsRel $\downarrow$}} & \multicolumn{2}{c||}{\textbf{Delta1 $\uparrow$}} & \multicolumn{2}{c||}{\textbf{AbsRel $\downarrow$}} & \multicolumn{2}{c||}{\textbf{Delta1 $\uparrow$}} & \multicolumn{2}{c||}{\textbf{AbsRel $\downarrow$}} & \multicolumn{2}{c||}{\textbf{Delta1 $\uparrow$}} & \multicolumn{2}{c||}{\textbf{AbsRel $\downarrow$}} & \multicolumn{2}{c|}{\textbf{Delta1 $\uparrow$}}                                                                                                                                                                                                                                                                                                                                 \\
    \cline{2-17}
    \multicolumn{1}{|c|}{}     & \textbf{\makecell{ GT}}                            & \textbf{VINS}                                    & \textbf{\makecell{ GT}}                            & \textbf{VINS}                                    & \textbf{\makecell{ GT}}                            & \textbf{VINS}                                    & \textbf{\makecell{ GT}}                            & \textbf{VINS}                                   & \textbf{\makecell{ GT}}               & \textbf{VINS}                         & \textbf{\makecell{ GT}}               & \textbf{VINS}                         & \textbf{\makecell{ GT}}               & \textbf{VINS}                         & \textbf{\makecell{ GT}}               & \textbf{VINS}                         \\
    \hline
    deg 1 poly                 & \textbf{\textcolor{darkgreen}{0.078}}              & \textbf{\textcolor{darkgreen}{0.124}}            & \textbf{\textcolor{darkgreen}{0.949}}              & \textbf{\textcolor{darkgreen}{0.855}}            & 0.080                                              & 0.151                                            & 0.937                                              & \textbf{\textcolor{darkgreen}{0.883}}           & 2.613                                 & 0.894                                 & 0.682                                 & 0.532                                 & 1.081                                 & 0.436                                 & 0.840                                 & 0.735                                 \\
    \hline
    deg 2 poly                 & 0.177                                              & 0.180                                            & \textbf{\textcolor{orange}{0.947}}                 & \textbf{\textcolor{orange}{0.851}}               & \textbf{\textcolor{orange}{0.073}}                 & 0.142                                            & 0.947                                              & \textbf{\textcolor{orange}{0.882}}              & 0.738                                 & 0.361                                 & 0.734                                 & 0.562                                 & 0.370                                 & 0.241                                 & 0.863                                 & 0.745                                 \\
    \hline
    deg 3 poly                 & 0.164                                              & 0.287                                            & 0.935                                              & 0.825                                            & 0.157                                              & 0.257                                            & \textbf{\textcolor{orange}{0.949}}                 & 0.861                                           & 20.259                                & 0.457                                 & 0.850                                 & 0.570                                 & 8.105                                 & 0.346                                 & 0.905                                 & 0.734                                 \\
    \hline
    deg 4 poly                 & 0.255                                              & 0.371                                            & 0.922                                              & 0.810                                            & 0.135                                              & 0.265                                            & 0.944                                              & 0.846                                           & \textbf{\textcolor{orange}{0.144}}    & 0.428                                 & \textbf{\textcolor{darkgreen}{0.890}} & 0.562                                 & \textbf{\textcolor{orange}{0.178}}    & 0.364                                 & \textbf{\textcolor{orange}{0.915}}    & 0.722                                 \\
    \hline
    deg 5 poly                 & 0.351                                              & 0.374                                            & 0.905                                              & 0.790                                            & 0.207                                              & 1.088                                            & 0.938                                              & 0.828                                           & 0.448                                 & 0.500                                 & \textbf{\textcolor{darkgreen}{0.890}} & 0.549                                 & 0.350                                 & 0.620                                 & 0.908                                 & 0.705                                 \\
    \hline
    exponential                & 0.177                                              & 0.167                                            & 0.735                                              & 0.732                                            & 0.136                                              & 0.143                                            & 0.817                                              & 0.786                                           & 0.246                                 & \textbf{\textcolor{orange}{0.341}}    & 0.566                                 & 0.447                                 & 0.193                                 & \textbf{\textcolor{orange}{0.229}}    & 0.691                                 & 0.634                                 \\
    \hline
    smoothing spline           & \textbf{\textcolor{orange}{0.092}}                 & 0.136                                            & 0.932                                              & 0.840                                            & 0.074                                              & \textbf{\textcolor{orange}{0.124}}               & 0.948                                              & 0.868                                           & 0.407                                 & 0.407                                 & 0.784                                 & \textbf{\textcolor{darkgreen}{0.608}} & 0.212                                 & 0.240                                 & 0.878                                 & \textbf{\textcolor{orange}{0.756}}    \\
    \hline
    % natural spline & 0.240 & 0.451 & 0.886 & 0.772 & 0.173 & 0.251 & 0.925 & 0.798 & 0.571 & 0.809 & 0.763 & 0.507 & 0.352 & 0.537 & 0.848 & 0.674\\
    % \hline
    monotonic smoothing spline & 0.094                                              & \textbf{\textcolor{orange}{0.128}}               & 0.927                                              & 0.847                                            & 0.076                                              & \textbf{\textcolor{darkgreen}{0.109}}            & 0.944                                              & 0.880                                           & 0.382                                 & 0.446                                 & 0.716                                 & \textbf{\textcolor{orange}{0.607}}    & 0.203                                 & 0.248                                 & 0.848                                 & \textbf{\textcolor{darkgreen}{0.761}} \\
    \hline
    monotonic spline           & 0.094                                              & 0.136                                            & 0.930                                              & 0.833                                            & \textbf{\textcolor{darkgreen}{0.069}}              & 0.132                                            & \textbf{\textcolor{darkgreen}{0.956}}              & 0.850                                           & \textbf{\textcolor{darkgreen}{0.116}} & \textbf{\textcolor{darkgreen}{0.264}} & \textbf{\textcolor{orange}{0.879}}    & 0.606                                 & \textbf{\textcolor{darkgreen}{0.096}} & \textbf{\textcolor{darkgreen}{0.185}} & \textbf{\textcolor{darkgreen}{0.917}} & 0.748                                 \\
    \hline
  \end{tabular}
  \caption{\label{tab:depthany_compare}Metric depth benchmarking results using the
    proposed rescaling methodologies and relative depth from DepthAnythingV2.
    The best performing approach is colored in green and the second best
    performing approach is colored in orange. Results for both the proposed approach, which
      rescales the depth using the 3D feature map from VINS, as well as ground truth (labeled
      as \textbf{GT}) are provided. The monotonic spline yields
    competitive performance in the confined space environments (i.e., \emph{mine}
    and \emph{sewer}) and also yields superior performance in the open environment
    (i.e., \emph{drone dome}).  Therefore, we leverage this approach for hardware
    experimentation.}
  \vspace{-0.15cm}
\end{table*}

\begin{table*}
  \setlength{\arrayrulewidth}{0.7pt}
  \renewcommand{\arraystretch}{1.3}
  \setlength{\tabcolsep}{1pt}
  \centering
  \captionsetup{font=small}\setlength{\arrayrulewidth}{0.7pt}\renewcommand{\arraystretch}{1.3}\setlength{\tabcolsep}{5pt}\centering
  \begin{tabular}{|>{\bfseries}l|c||c||c||c||c||c||c||c||c||c||c||c|}\hline\multicolumn{1}{|c|}{\textbf{Model}} & \multicolumn{3}{c||}{\textbf{Mine}}           & \multicolumn{3}{c||}{\textbf{Sewer}} & \multicolumn{3}{c||}{\textbf{Drone Dome}} & \multicolumn{3}{c|}{\textbf{Weighted Average}}                                                                                                                                                \\
               \cline{2-13}\multicolumn{1}{|c|}{}                                                             & \multicolumn{1}{c||}{\textbf{\makecell{AbsRel                                                                                                                                                                                                                                                                                    \\$\downarrow$}}} & \multicolumn{1}{c||}{\textbf{\makecell{Delta1\\$\uparrow$}}} & \multicolumn{1}{c||}{\textbf{\makecell{FPS\\$\uparrow$}}} & \multicolumn{1}{c||}{\textbf{\makecell{AbsRel\\$\downarrow$}}} & \multicolumn{1}{c||}{\textbf{\makecell{Delta1\\$\uparrow$}}} & \multicolumn{1}{c||}{\textbf{\makecell{FPS\\$\uparrow$}}} & \multicolumn{1}{c||}{\textbf{\makecell{AbsRel\\$\downarrow$}}} & \multicolumn{1}{c||}{\textbf{\makecell{Delta1\\$\uparrow$}}} & \multicolumn{1}{c||}{\textbf{\makecell{FPS\\$\uparrow$}}} & \multicolumn{1}{c||}{\textbf{\makecell{AbsRel\\$\downarrow$}}} & \multicolumn{1}{c||}{\textbf{\makecell{Delta1\\$\uparrow$}}} & \multicolumn{1}{c|}{\textbf{\makecell{FPS\\$\uparrow$}}} \\
               \hline
               Without TensorRT                                                                               & {\textbf{0.094}}                              & {\textbf{0.930}}                     & 5                                         & {\textbf{0.069}}                               & {\textbf{0.956}} & 5             & {\textbf{0.117}} & {\textbf{0.879}} & 5             & {\textbf{0.096}} & {\textbf{0.917}} & 5             \\
               \hline
               With TensorRT                                                                                  & 0.096                                         & 0.929                                & {\textbf{20}}                             & 0.070                                          & 0.949            & {\textbf{20}} & 0.121            & 0.869            & {\textbf{19}} & 0.099            & 0.911            & {\textbf{20}} \\
               \hline
  \end{tabular}
  \caption{\label{tab:tensor_rt_compare} Inference speed and accuracy with and without TensorRT using
    monotonic spline rescaling and DepthAnythingV2 on-board an Orin AGX. The key takeaway is the
    TensorRT model suffers negligible performance degradation while substantially increasing
    the frame rate for depth prediction. Therefore, we leverage the TensorRT model for
    hardware experimentation.}
  \vspace{-0.25cm}
\end{table*}

We conduct ablation studies and evaluate results using Absolute Relative Error
and $\delta_1$ on simulated datasets. For these studies, we employ
DepthAnythingV2~\cite{yang2024depthv2} as our monocular relative depth
estimation (MDE) model. DepthAnythingV2 is included given its prior use
by~\citet{saviolo2024reactive}.  To enable real-time operation on constrained
quadrotor hardware, we use the DepthAnythingV2 small model with $24.8$~M
parameters, which strikes a balance between accuracy and inference speed.

Although the DepthAnything model~\cite{yang2024depth} suggests fine-tuning for a
target environment to achieve accurate metric depth, this approach is infeasible
for robotic systems operating in unknown surroundings. We use the default
parameter settings given by the authors of each model, including the
pre-processing steps of each RGB image, while accounting for our RGB camera
intrinsics. We conduct these evaluations on a desktop with Ubuntu 20.04 (ROS1
Noetic), an Intel i9-14900K processor, an NVIDIA RTX 4090 GPU, and 32 GB of RAM.

\begin{figure}
  \centering
  \includegraphics[width=0.98\linewidth,trim=10 32 10 21,clip]{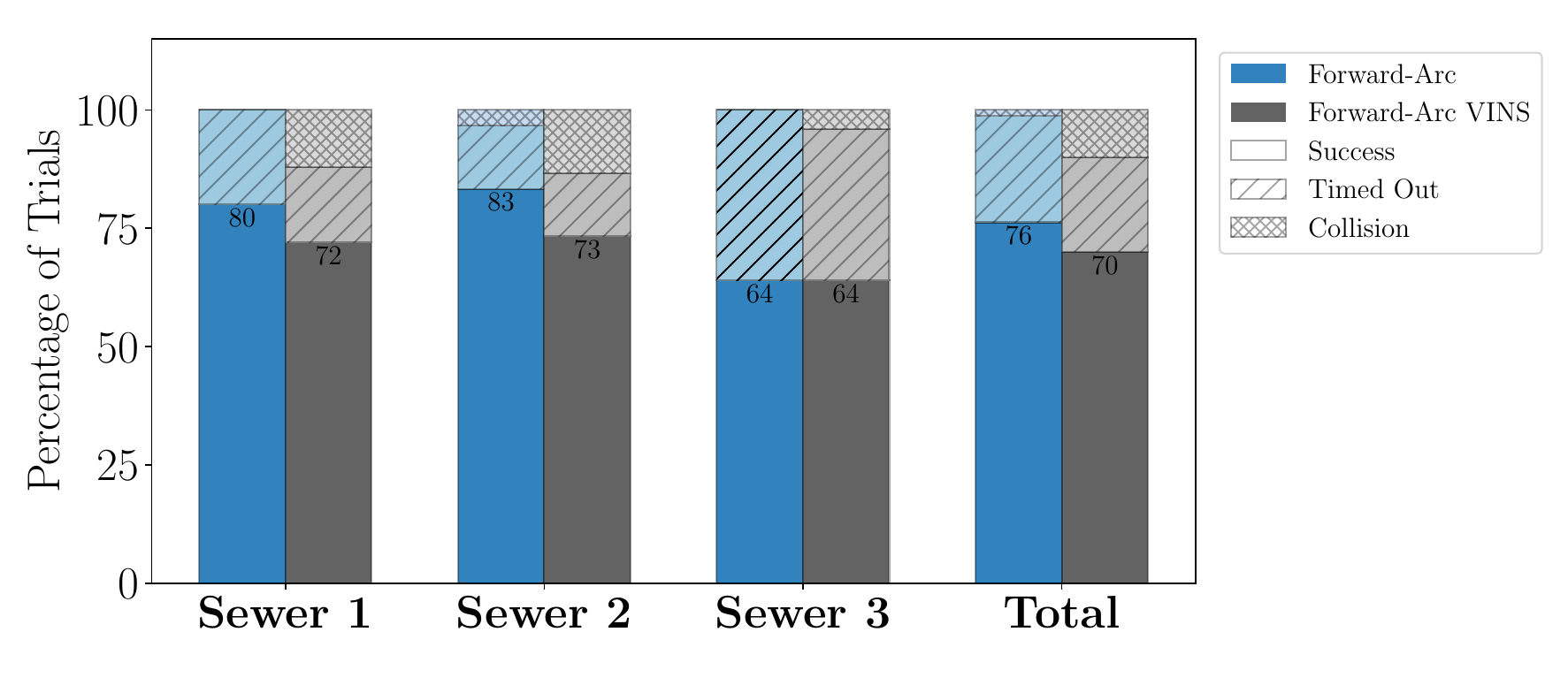}
  \caption{\label{fig:agx_benchmark}Performance comparison
    of autonomous navigation in the simulated sewer environments using our rescaled
    metric depth estimation approach (shown in grey) and the depth camera data (shown in blue).
    The proposed approach suffers minor degradation compared to the depth camera image,
    which is expected as scale is estimated using the fused monocular camera and IMU
    data.}
  \vspace{-0.3cm}
\end{figure}

\Cref{tab:depthany_compare} shows the result for the proposed rescaling methods,
when used with relative depth estimates from DepthAnythingV2. The rescaling
strategies are compared using ground truth depth from the simulated depth camera
image queried at the pixel locations corresponding to the sparse 3D points from
the VINS projected into the image plane. The columns labeled \textbf{GT}
leverage the ground truth depth data to perform rescaling and provide a baseline
comparison for our approach. The \textbf{GT} columns are expected to outperform
our approach, because our approach does not use ground truth data. Sparse
features in our approach are optimized over a sliding-window of frames, where
corner detection introduces errors over time. Since the 3D sparse feature
locations rely on corner detection accuracy, these accumulated tracking errors
can propagate across the optimization window, leading to degradation in
accuracy. Nonetheless, the key takeaway is that using the sparse 3D feature
map from VINS yields comparable performance for estimating depth and this is
usable for autonomous navigation.

From the weighted average of evaluation metrics across the environments,
we observe the monotonic spline rescaling strategy yields the best performance
across both confined and open space environments. Because we cannot make any
assumptions about the dataset or the shape of the relative-metric disparity
distribution and the monotonic spline rescaling strategy performs well across
both confined and open space environments, this is the approach we use for the
remaining experiments.
\begin{figure*}
  \centering
  \includegraphics[width=0.32\linewidth]{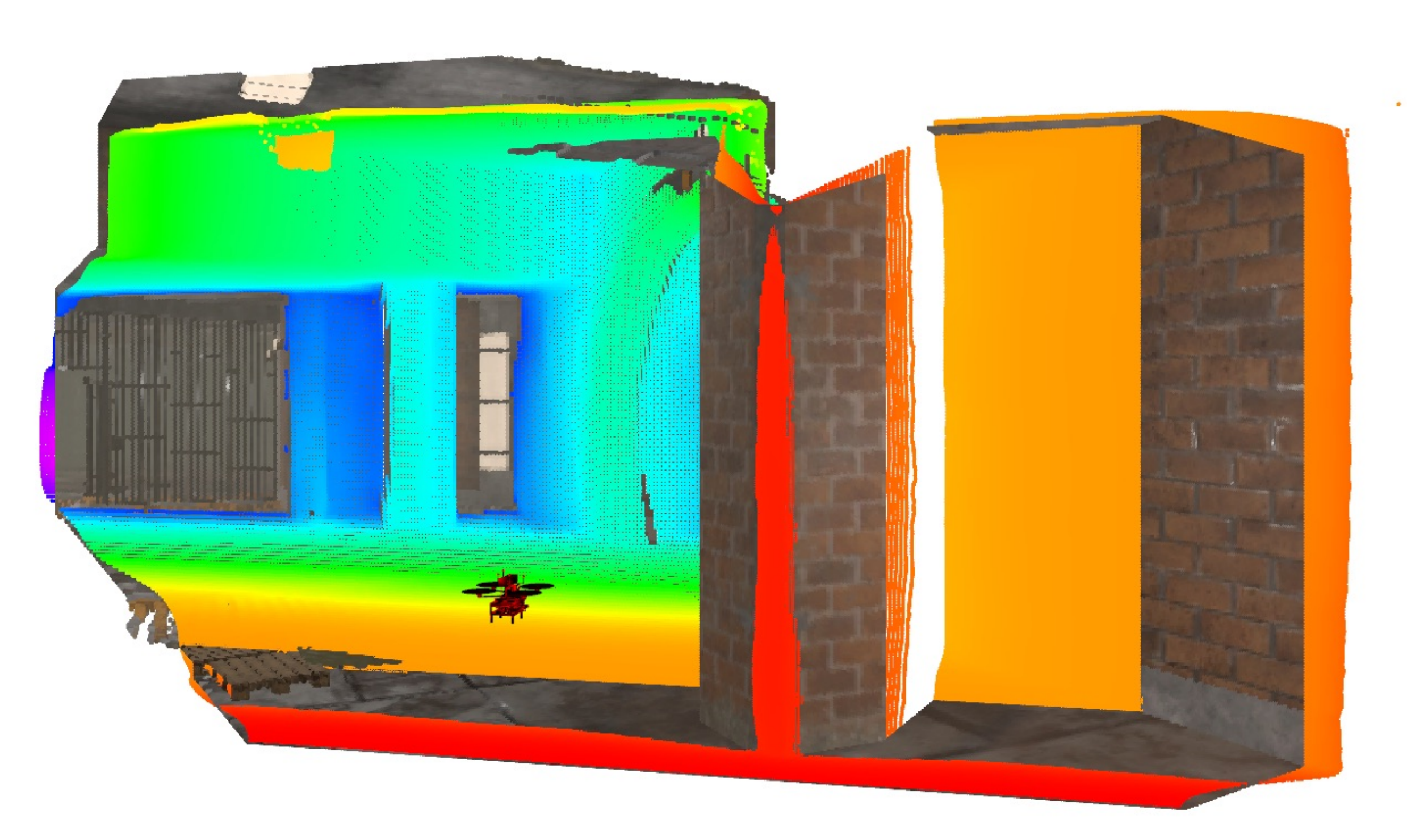}
  \includegraphics[width=0.32\linewidth]{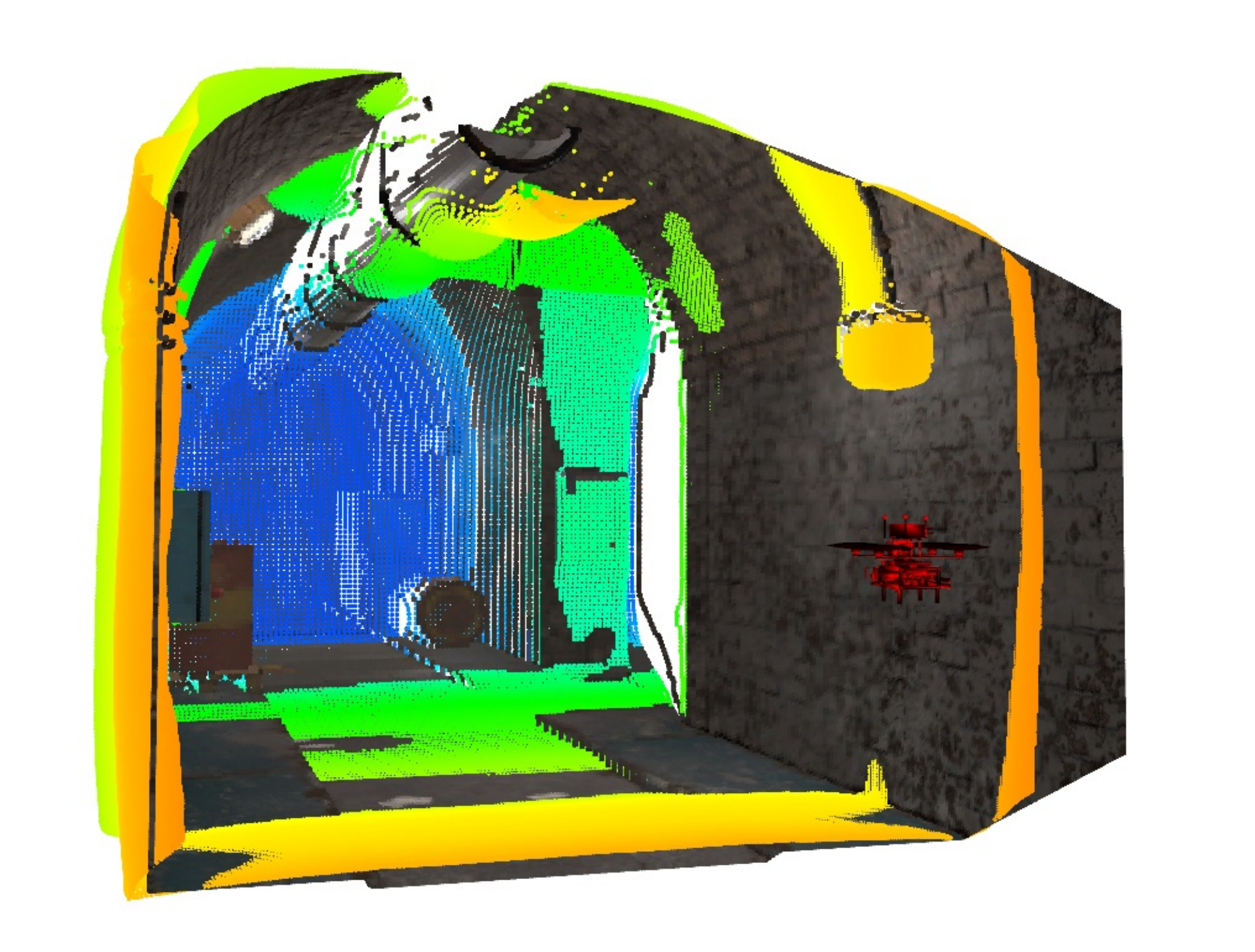}
  \includegraphics[width=0.32\linewidth]{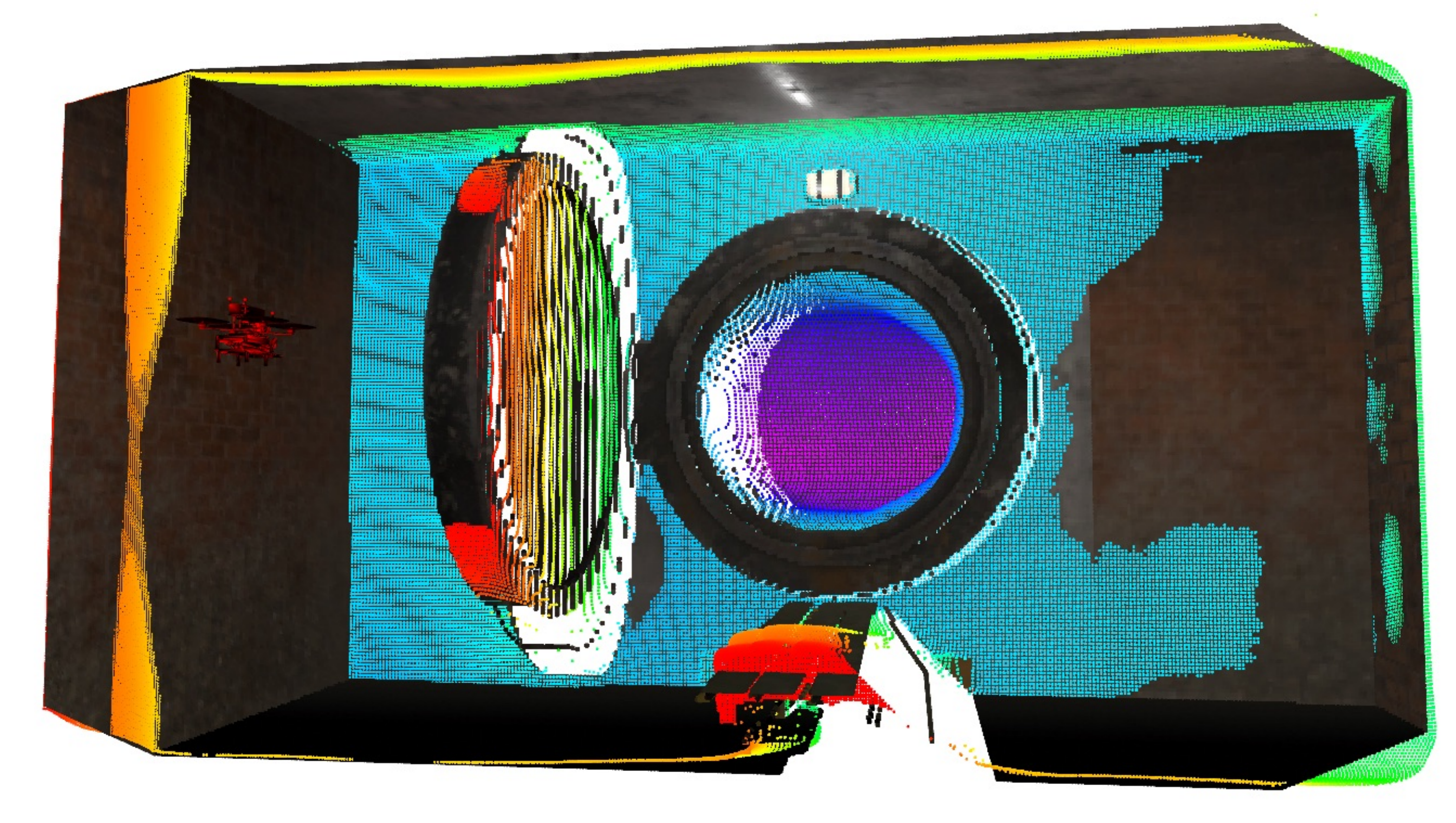}
  \caption{Representative simulated scenes used to evaluated
    the proposed approach. The colors ranging from red (closer) to purple (further away)
    represent the metric depth estimates after rescaling relative to the robots
    position (shown as a red quadrotor). The predicted depth values closely
    align with the ground truth, demonstrating the accuracy of the
    methodology.}
  \label{fig:sim_nav_examples}
  \vspace{-0.3cm}
\end{figure*}

\begin{figure}
  \centering

  \subfloat[\label{sfig:hardware_test_3pv}]{\includegraphics[width=\linewidth]{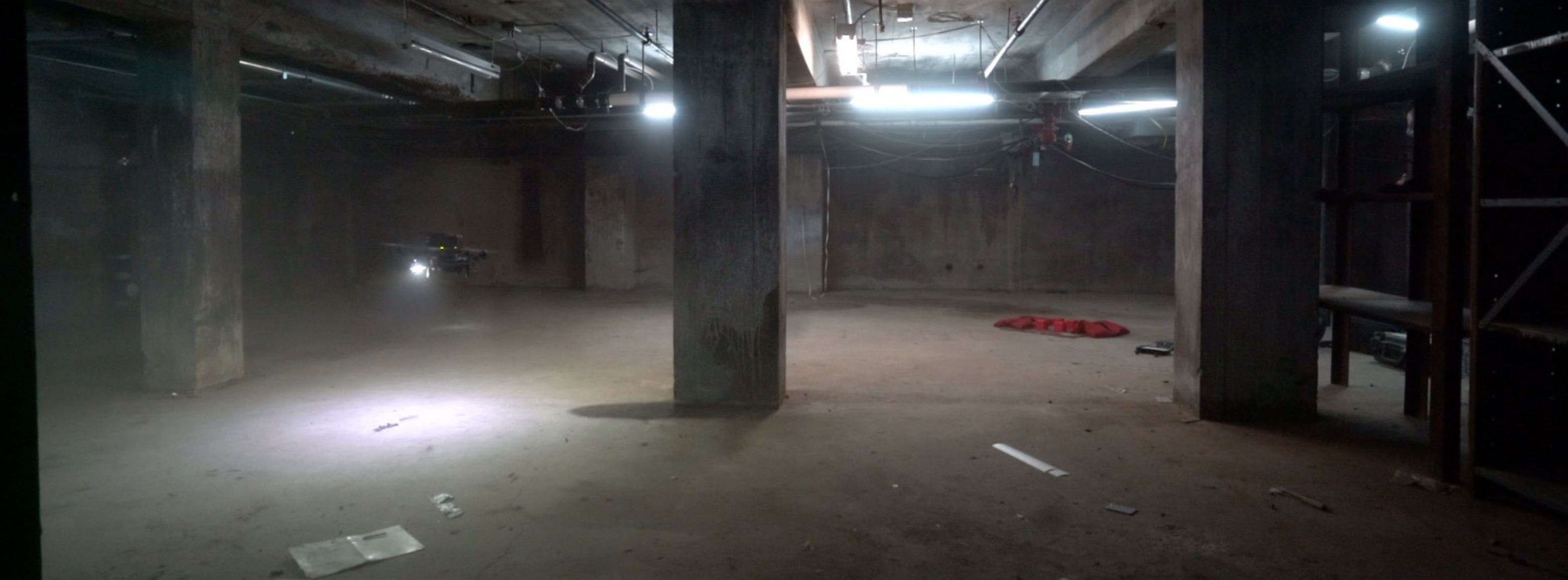}}\\
  \subfloat[\label{sfig:hardware_test_feature}]{\includegraphics[width=0.33\linewidth]{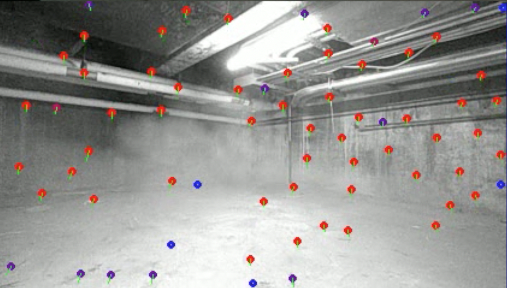}}\hfill
  \subfloat[\label{sfig:hardware_test_realsense}]{\includegraphics[width=0.32\linewidth]{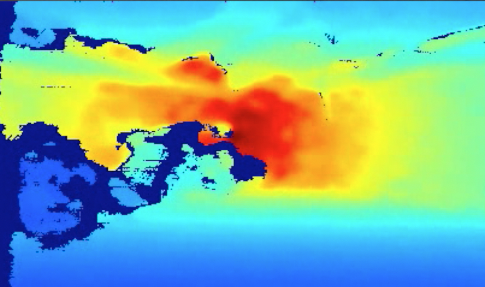}}\hfill%
  \subfloat[\label{sfig:hardware_test_proposed}]{\includegraphics[width=0.33\linewidth]{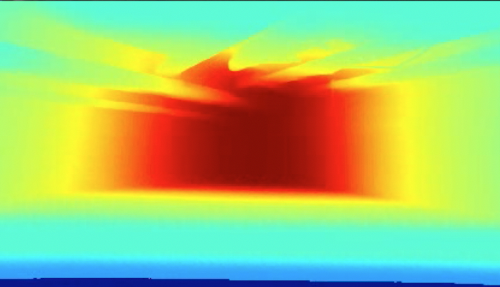}}

  \caption{\label{fig:hardware_catacomb}
  Images and data from one of the hardware experiments.
    \protect\subref{sfig:hardware_test_3pv} provides a third-person view of the robot navigating a dusty industrial tunnel environment.
    \protect\subref{sfig:hardware_test_feature} illustrates the corner detections.
    \protect\subref{sfig:hardware_test_realsense} illustrates how dust affects the active stereo depth image from the RealSense.
    \protect\subref{sfig:hardware_test_proposed} provides the results from our proposed method.}
  \vspace{-0.3cm}
\end{figure}

\begin{figure}
  \centering
  \subfloat[\label{sfig:tripod_rgb}]{\includegraphics[width=0.33\linewidth]{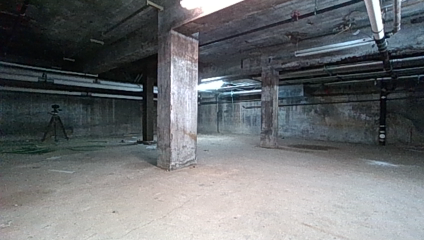}}\hfill
  \subfloat[\label{sfig:hardware_test_realsense_no_tripod}]{\includegraphics[width=0.32\linewidth]{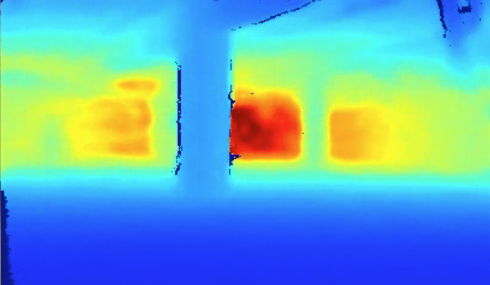}}\hfill
  \subfloat[\label{sfig:hardware_test_tripod}]{\includegraphics[width=0.33\linewidth]{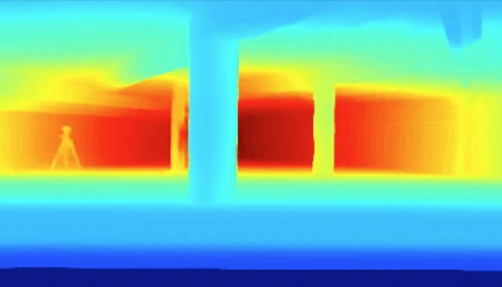}}

  \caption{Comparison of environmental details during hardware experiments.
    \protect\subref{sfig:tripod_rgb} RGB image from the RealSense color camera
    \protect\subref{sfig:hardware_test_realsense_no_tripod} provides a view of the depth image from the RealSense.
    \protect\subref{sfig:hardware_test_tripod} provides the results from our proposed method.
    \label{fig:hardware_tripod_compare}
  }
    \vspace{-0.3cm}
\end{figure}

\subsection{Autonomous Aerial Navigation in Simulation}
The proposed metric depth estimation approach is integrated with the
motion primitives-based planner from~\citep{lee2024rapid} to enable autonomous
aerial navigation. We choose the simulated \textit{sewer} environment
for benchmarking navigation performance.  We execute the planner at a max
velocity of \SI{2}{\meter\per\second}, with a trajectory duration of
\SI{2}{\second}, and a robot radius of \SI{1}{\meter} to mitigate potential
collisions caused by noise in the estimated depth maps. The
Flightmare~\citep{song2021flightmare} simulation ran on a host PC equipped with
an AMD Ryzen 9950X3D and an RTX 5090 GPU. The proposed depth estimation method,
VINS, and the motion planner are executed using a NVIDIA Jetson Orin AGX connected
to the host computer via a local area network.
Each environment has 5 to 6 start and goal pairs, each pair is run
for $5$ trials with $120$ second time out cutoff.

To obtain higher inference rates, we convert the model into a TensorRT format
with FP32 precision and perform benchmarking.~\Cref{tab:tensor_rt_compare}
presents the trade-off between accuracy and inference speed onboard the NVIDIA
Jetson Orin AGX. The results show only a minor accuracy drop, accompanied by a
substantial boost in FPS performance.

Quantitative results for navigation performance are shown
in~\cref{fig:agx_benchmark}. We evaluate the impact of the proposed depth
prediction method on (1) the number of collisions incurred and (2) the number of
times the robot was able to reach the goal. The \textit{Forward-Arc} case
in~\cref{fig:agx_benchmark} uses ground-truth depth data for navigation while
the proposed method is labeled \textit{Forward-Arc VINS}. Navigation using
estimated depth yields reliable performance with only minimal degradation
compared to using ground truth depth. Qualitatively, the predicted point clouds
align with the ground truth point clouds (\cref{fig:sim_nav_examples}). The
number of timeouts and collision rates are comparable between the two methods,
demonstrating that the predicted metric depth effectively enables autonomous
navigation using a single RGB camera and an IMU instead of the stereo camera used
in~\citep{saviolo2024reactive,lee2024rapid}.

\subsection{Real World Deployment\label{ssec:real-world}}
The approach was deployed on a custom-built quadrotor aerial robot
equipped with a NVIDIA Orin AGX (\SI{32}{\giga\byte}). The forward-facing
global shutter RGB camera is from a RealSense D455 sensor. The IMU data comes
from the flight controller, which uses a custom version of the PX4 firmware
running on an mRo Pixracer. During hardware testing, we ran two VINS instances:
one for the depth estimation from the forward-facing camera and another for state
estimation using a downward-facing Matrix Vision BlueFox global shutter grayscale
camera. All autonomy functions, including obstacle avoidance were run onboard
the Orin AGX. Consequently, the load on the CPU was high, which
led to slightly reduced depth-estimation inference rate at 15 FPS. This
performance remained compatible with system requirements, since the planner
operates at 12 Hz, and no systematic failures were observed.

Experiments were conducted in a constrained, dusty catacomb-like
environment with concrete pillars (\cref{sfig:hardware_test_3pv}). The
target location was positioned 7 meters from the robot's starting
point, behind several pillars, to evaluate obstacle-occluded
navigation performance. The robot successfully completed two
navigation trials without prior knowledge of the environment.  The
system demonstrated robustness under dust conditions
(\cref{fig:hardware_catacomb}).  We also observed that our depth
estimation method captures fine details. For example, we were able to
detect the tripod in the depth estimation output, as shown in
\cref{fig:hardware_tripod_compare}.

\section{LIMITATIONS}
Despite successful navigation, several limitations remain. The proposed depth
estimation performance is expected to degrade in open scenes where large
portions of the image consist of sky views due to the lack of features or when
prominent foreground objects abruptly disappear from the scene. These situations
result in degraded metric depth estimates; however, recovery is typically
achieved once forward-facing feature points are re-established.  Because there
may be chattering in the depth predictions from one frame to the next, planning
approaches that select actions close to surfaces (e.g., hug a wall to make a
right-hand turn to get to the goal) may be at greater risk of collision.  To
mitigate for this, a larger collision radius should be used or planning
approaches that include gradient information to enable the vehicle to stay far
away from obstacles.  Furthermore, the method relies heavily on the accuracy of
sparse feature points. Depth estimation becomes unreliable when predictions are
made outside the depth range represented by the sampled sparse features.

%%% Local Variables:
%%% mode: latex
%%% TeX-master: "../main"
%%% End:

\section{CONCLUSION}
In this work, we presented a methodology to predict metric depth from
monocular RGB images and an inertial measurement unit (IMU)
The approach feeds an RGB image to a monocular
depth estimation network without fine-tuning and rescales the inferred depth
image using the sparse 3D feature points from a sliding window visual-inertial
navigation system. We evaluate several rescaling strategies using diverse
simulation data. Our results indicate that DepthAnythingV2 with monotonic
splines achieves the most consistent output in both open and confined spaces.
Finally, we deploy that pipeline to run navigation in simulation and hardware,
validating the depth estimation's robustness through reliable obstacle avoidance
without reliance on depth sensors or prior environment knowledge.

Potential directions for future research include optimizing rescaling algorithms
to support higher-rate metric depth predictions for faster navigation, improving
the accuracy and spread of VINS sparse features, developing uncertainty-aware
weighted rescaling methods and uncertainty-aware planning strategies (e.g.,
favoring translational motion when depth predictions or sparse features exhibit
low confidence).

%%% Local Variables:
%%% mode: latex
%%% TeX-master: "../main"
%%% End:

\ifthenelse{\equal{\arxivmode}{false}}{
}{
  \section*{ACKNOWLEDGEMENTS}
The authors would like to thank Jonathan Lee for valuable
discussions and insights. This material is based upon work supported
in part by the Army Research Laboratory and the Army Research Office
under contract/grant number W911NF-25-2-0153.
}

{
  \balance
  \footnotesize
  {
    \bibliographystyle{IEEEtranN}
    \bibliography{refs}
  }
}

\end{document}